\documentclass[sigconf]{acmart}

\AtBeginDocument{%
  }

\setcopyright{acmlicensed}
\copyrightyear{2026}
\acmYear{2026}
\setcopyright{cc}
\setcctype{by}
\acmConference[MM '26]{Proceedings of the 34th ACM International Conference on Multimedia}{November 10--14, 2026}{Rio de Janeiro, Brazil}
\acmBooktitle{Proceedings of the 34th ACM International Conference on Multimedia (MM '26), November 10--14, 2026, Rio de Janeiro, Brazil}
\acmDOI{10.1145/3767308.3836346}
\acmISBN{979-8-4007-2213-4/2026/11}

\usepackage{multirow}

\begin{document}

\title{ForceU-VLA: A Force-Aware Vision–Language–Action Model for Embodied Ultrasound Scanning}

\author{Xingzheng Wu}
\authornote{Both authors contributed equally to this research.}
\affiliation{%
  \institution{Faculty of Computer Science and Technology, Ocean University of China}
  \city{Qingdao}
  \country{China}}
\email{wuxingzheng@stu.ouc.edu.cn}

\author{Cheng Zhang}
\authornotemark[1]
\affiliation{%
  \institution{Faculty of Computer Science and Technology, Ocean University of China}
  \city{Qingdao}
  \country{China}}
\email{zhangcheng@stu.ouc.edu.cn}

\author{Guihao Yan}
\affiliation{%
  \institution{Faculty of Computer Science and Technology, Ocean University of China}
  \city{Qingdao}
  \country{China}}
\email{yanguihao@stu.ouc.edu.cn}

\author{Xifeng Hu}
\affiliation{%
  \institution{School of Information Science and Engineering, Shandong University}
  \city{Qingdao}
  \country{China}}
\email{202220466@mail.sdu.edu.cn}

\author{Zhi Liu}
\correspondingauthor
\affiliation{%
  \institution{School of Information Science and Engineering, Shandong University}
  \city{Qingdao}
  \country{China}}
\email{liuzhi@sdu.edu.cn}

\author{Qing Cai}
\correspondingauthor
\affiliation{%
  \institution{Innovation School of Artificial Intelligence, Hefei University of Technology}
  \city{Hefei}
  \country{China}}
\email{caiqing@hfut.edu.cn}

\renewcommand{\shortauthors}{Xingzheng Wu et al.}

\begin{abstract}
Embodied intelligent ultrasound scanning enables the automation and standardization of the ultrasound examination process by integrating perception, decision-making, and execution capabilities. However, existing methods suffer from loosely coupled modeling between force and ultrasound modalities and lack awareness of scanning stages, which limits their ability to capture dynamic probe–tissue interactions. To address these issues, we propose ForceU-VLA, a force-aware Vision–Language–Action model for autonomous embodied ultrasound scanning, which leverages force signals and ultrasound image feedback throughout the scanning process to enable accurate and high-quality ultrasound acquisition. Firstly, we propose a Force-Ultrasound Synergistic Fusion Module (FUSFM) that synergistically fuses ultrasound visual and force-feedback information to provide stable, reliable guidance for probe motion. Secondly, a Stage-Adaptive Modulation Mechanism (SAMM) is proposed to accommodate the task requirements across different scanning stages by adaptively modulating multimodal features to enhance their representation quality. Additionally, we introduce ForceU-VLA-Data, a real-world, force-aware embodied ultrasound dataset that integrates visual, force, and action signals, including data from two organs across five representative clinical scanning views, and comprising 450 expert-collected trajectories with approximately 100,000 synchronized multimodal frames. Extensive experimental results demonstrate that ForceU-VLA significantly improves contact stability and probe pressure regulation in embodied ultrasound scanning, thereby effectively enhancing task execution quality and overall system reliability. The source code is available at https://github.com/VMVLab/ForceU-VLA.
\end{abstract}



\maketitle

\section{Introduction}

Ultrasound imaging has been widely adopted in clinical diagnosis due to its advantages of safety, real-time capability, and cost-effectiveness \cite{bib1,bib2,bib37}. However, ultrasound scanning is highly operator-dependent, which necessitates continuous and precise adjustment of the probe’s position, orientation, and contact force to reliably acquire diagnostically meaningful and clinically relevant standard views \cite{bib48}. In recent years, embodied ultrasound scanning systems have increasingly enabled automated and standardized probe manipulation through the tight integration of environmental perception, decision-making, and physical interaction \cite{bib47}, thereby effectively reducing reliance on operator expertise and enhancing the overall accessibility and consistency of diagnostic services in routine clinical practice settings \cite{bib3,bib4,bib5,bib6}.

\begin{figure}[t]
\centering
\includegraphics[width=0.47\textwidth]{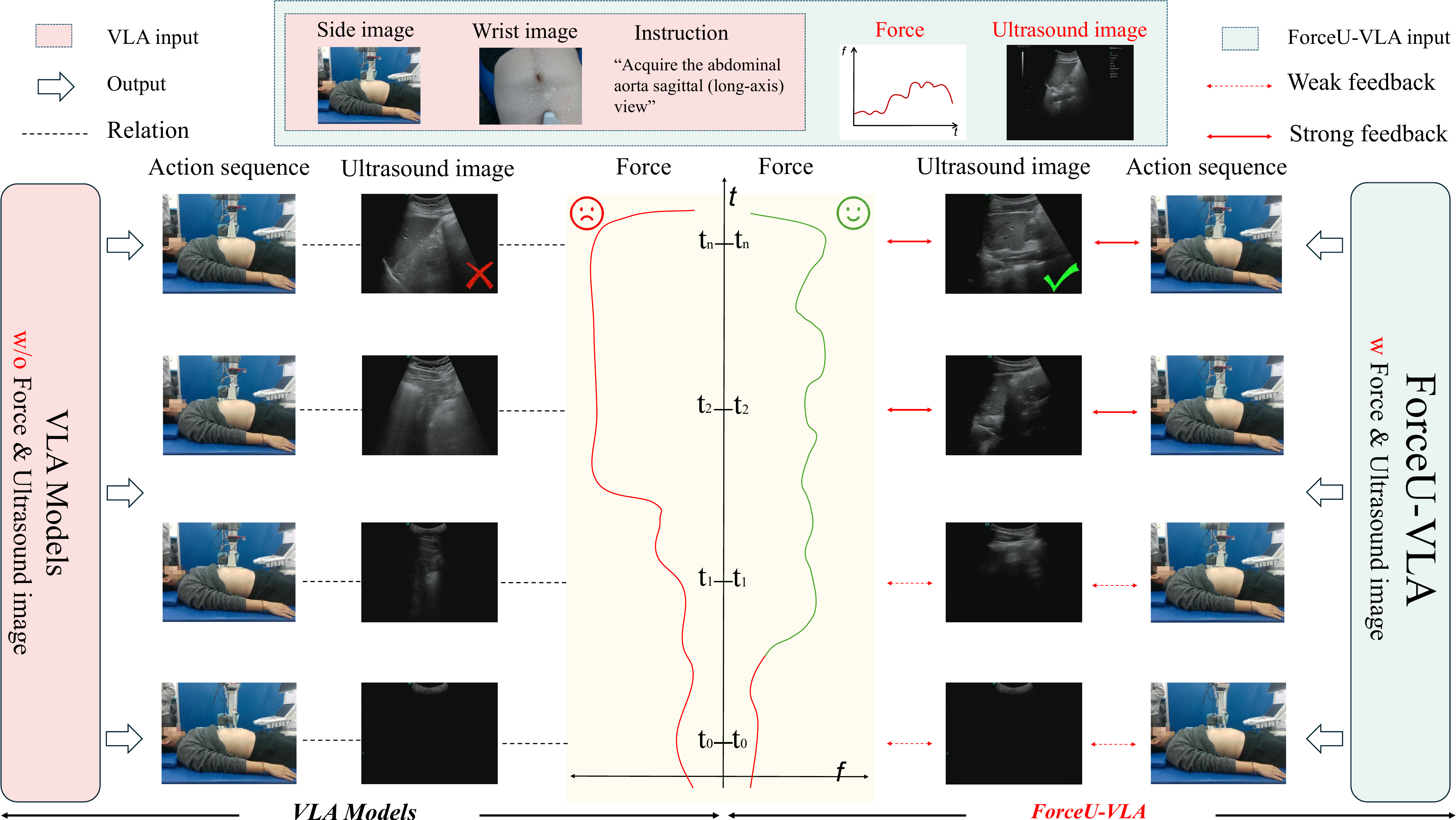}
\caption{Comparison between classical VLA methods and the proposed ForceU-VLA. By explicitly modeling both ultrasound and force information, ForceU-VLA improves contact stability and probe pressure regulation, thereby enhancing both the efficiency and quality of ultrasound scanning.}
\label{fig1}
\end{figure}

During embodied intelligence–assisted ultrasound scanning, force information plays a crucial role in maintaining stable contact, ensuring interaction safety, and enhancing the robustness of task execution \cite{bib7,bib39,bib40,bib41}. Existing studies on the utilization of force information can be broadly categorized into three groups. Force-control-based methods achieve stable interaction between the probe and tissue by constructing six-dimensional force regulation or hybrid control frameworks \cite{bib8,bib9,bib10,bib11}. However, such approaches primarily focus on low-level contact control and have yet to incorporate force information into higher-level perception and decision-making processes. Task-driven force–motion coordination-based methods establish mappings between contact force and tissue deformation to enable feedforward correction of scanning trajectories \cite{bib12}. By integrating path planning with interaction control mechanisms, these methods support stable scanning and anomaly recovery in complex task scenarios \cite{bib13,bib14,bib15}. Nevertheless, it is typically treated as an auxiliary variable for control and planning, rather than being embedded within a unified perception–decision modeling framework. Multimodal fusion and learning-based methods leverage deep reinforcement learning to integrate visual and tactile information for end-to-end scanning policy generation \cite{bib16}, and further enhance system autonomy through reinforcement learning or large-scale imitation learning \cite{bib17,bib18}. Despite these initial efforts to incorporate force information into perception and decision-making, a unified modeling framework is still lacking, and the complex interactions among force, vision, and action remain insufficiently characterized.

In summary, although existing approaches have explored the use of force information from the perspectives of contact control, task planning, and multimodal learning, two key limitations remain. First, the coupling between force information and ultrasound imaging is often loosely established, lacking systematic modeling of their temporal alignment and coordinated feedback mechanisms. This limitation hinders the ability to fully capture the dynamic interplay between haptic and visual signals during probe–tissue interaction. Second, ultrasound scanning is inherently a contact-rich process with distinct operational stages. Different phases, such as pre-contact, near contact, and stable contact, exhibit substantial differences in both perceptual characteristics and control requirements. However, most existing methods adopt unified feature representations and decision-making strategies, without incorporating stage-aware adaptive mechanisms to account for these variations.

To address these limitations, we propose a ForceU-VLA for force-aware modeling and closed-loop interaction control in embodied ultrasound scanning, as illustrated in Fig.~\ref{fig1}. Specifically, a Force-Ultrasound Synergistic Fusion Module (FUSFM) is proposed to jointly model ultrasound visual information and force feedback signals through cross-modal interaction, enabling effective multimodal fusion and providing stable and reliable feedback for probe motion. Furthermore, a Stage-Adaptive Modulation Mechanism (SAMM) is proposed to characterize feature discrepancies across scanning stages and enhance representation capability by adaptively modulating multimodal features. In addition, a ForceU-VLA-Data dataset is constructed, which is a real-world embodied ultrasound scanning dataset with force feedback. It covers two organs, namely the liver and kidney, and five standard scanning views, comprising 450 expert demonstration trajectories and approximately 100,000 synchronized time steps. Extensive experimental results consistently demonstrate that the proposed ForceU-VLA significantly improves contact stability and force regulation capability.

The contributions of this paper are summarized as follows:
\begin{itemize}
    \item We propose ForceU-VLA, a force-aware VLA framework for autonomous embodied ultrasound scanning, which leverages force signals and ultrasound image feedback in a coordinated manner throughout the scanning process to enable accurate and high-quality probe–tissue interactions.
    
    \item A real-world embodied ultrasound scanning dataset, ForceU-VLA-Data, is constructed. It covers two organs (liver and kidney) and five standard scanning views, comprising 450 expert demonstration trajectories and approximately 100,000 temporally synchronized samples.
    
    \item Extensive experimental results demonstrate that the proposed ForceU-VLA significantly outperforms existing methods in contact stability and force regulation, leading to substantially improved task completion quality and enhanced overall system reliability.
\end{itemize}

\begin{figure*}[t]
\centering
\includegraphics[width=0.95\textwidth]{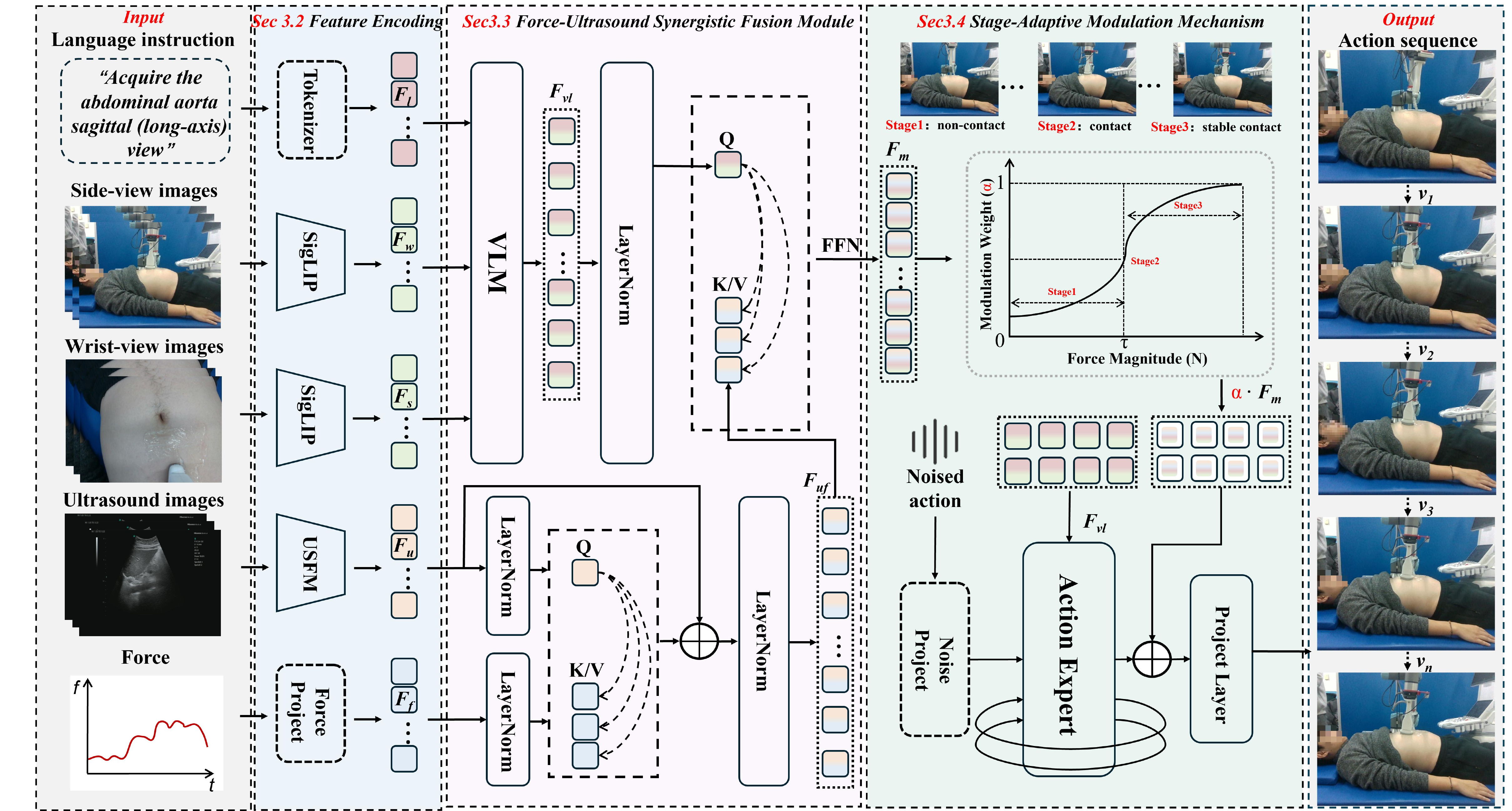}
\caption{Overview of ForceU-VLA. First, multimodal inputs are encoded into high-dimensional representations. Then, the Force–Ultrasound Synergistic Fusion Module performs collaborative fusion. Based on the fused features, the Stage-Adaptive Modulation Mechanism applies stage-aware adaptive weighting. Finally, the model generates the action sequence.}
\label{fig2}
\end{figure*}

\section{Related Works}
\subsection{Embodied Ultrasound Scanning systems}

Embodied ultrasound scanning systems integrate perception, decision-making, and action to enable autonomous, consistent, and high-quality ultrasound acquisition while reducing reliance on operator expertise \cite{bib19,bib20}. Existing embodied ultrasound scanning systems can be broadly categorized into reinforcement learning–based and imitation learning–based approaches. Reinforcement learning-based methods \cite{bib23,bib43,bib44,bib45} enable agents to acquire probe manipulation policies through trial-and-error interactions with the environment. They offer strong adaptability and autonomy, but often suffer from low sample efficiency, prolonged training processes, and potential safety concerns during exploration. In contrast, imitation learning-based methods \cite{bib21,bib22,bib42} leverage expert demonstrations to directly acquire scanning behaviors, resulting in faster convergence and more stable performance. However, these methods are constrained by the quality and diversity of demonstration data and often exhibit limited generalization to unseen scenarios. In addition, teleoperation-based systems \cite{bib24,bib25} have been explored to provide enhanced flexibility and enable access to medical expertise in resource-limited or geographically isolated settings. However, they still rely heavily on human operators and lack full autonomy. Despite these advances, existing methods exhibit fundamental limitations in robustness and generalization, which hinder their effective deployment across diverse patient anatomies and clinical environments, particularly under varying contact conditions and operator-independent settings.

\subsection{Vision–Language–Action Models with Force Feedback}

The incorporation of force information has significantly enhanced the physical interaction robustness and operational precision of VLA-based methods in contact-rich tasks \cite{bib46}, leading to more stable control and safer human–robot interaction outcomes. At the architectural level, ForceVLA \cite{bib26} and FAVLA \cite{bib27} elevate high-frequency force feedback to a primary modality by introducing a Mixture-of-Experts mechanism and a fast–slow decoupling structure, respectively, thereby enabling rapid responses to complex contact variations. In addition, Force Policy \cite{bib28} establishes a hierarchical control framework that combines global visual planning with local haptic correction, allowing the robot to dynamically adjust its manipulation state under external disturbances to maintain stable contact. From the perspective of training strategies and generalization capability, CRAFT \cite{bib29} integrates curriculum learning with an information bottleneck mechanism, encouraging the model to prioritize haptic signals during early training and effectively mitigating the issue of visual dominance in multimodal learning. FD-VLA \cite{bib30} further introduces a cross-modal force distillation module, which reconstructs haptic representations without relying on physical force sensors, thus significantly reducing deployment costs. Moreover, TER-DAgger \cite{bib31} incorporates a force-aware anomaly detection mechanism together with a human-in-the-loop data collection strategy, alleviating the covariate shift problem in imitation learning.

\section{Method}
\subsection{Overview}

As shown in Fig.~\ref{fig2}, ForceU-VLA processes language instructions, wrist- and side-view images, ultrasound images, and force signals through dedicated encoders. The extracted features are fused by FUSFM to model probe–tissue interactions and then adaptively modulated by SAMM according to the scanning stage. Finally, the stage-aware features and noise vectors are processed by the action expert to generate continuous robot actions for stable and precise autonomous ultrasound scanning.

\subsection{Multimodal Feature Encoding}

The input modalities are first encoded into unified feature representations. Language instructions are tokenized and transformed into sequential features $F_l$, providing high-level semantic guidance for the task. Wrist-view and side-view images are processed using SigLIP \cite{bib32}, producing visual features $F_w$ and $F_s$, which primarily capture spatial context and external environmental information, including the relative pose of the probe with respect to the human body and global and local interaction states. Given the distinct characteristics of ultrasound images compared to natural images, a universal US foundation model (USFM) \cite{bib33} is employed to extract ultrasound features $F_u$. Unlike external visual observations, $F_u$ encodes internal anatomical structures and tissue-specific information, which are crucial for identifying clinically relevant regions and guiding precise scanning. Force signals (6D, comprising 3D forces and 3D torques) are embedded via a linear force projection module into a 2048-dimensional representation to obtain $F_f$, matching the embedding dimension of the visual tokens. This representation captures the physical interaction between the probe and the human body, including contact state and pressure variations.

\subsection{Force-Ultrasound Synergistic Fusion Module (FUSFM)}

Efficient fusion of heterogeneous modalities remains a key challenge in multimodal modeling. Existing methods, such as ForceVLA, incorporate force signals into vision–language frameworks through Mixture-of-Experts fusion, but often treat them as auxiliary inputs, limiting the modeling of strongly coupled physical dynamics. To address this issue in embodied ultrasound scanning, we propose the Force–Ultrasound Synergistic Fusion Module (FUSFM), which jointly encodes force and ultrasound information and performs cross-modal alignment, thereby improving multimodal perception and decision-making.

First, the instruction features $F_l$ are concatenated with the visual features $F_w$ and $F_s$, and fed into a pretrained vision–language model (VLM) \cite{bib34} to obtain a global semantic representation.

\begin{equation}
F_{vl} = \mathrm{VLM}([F_l; F_w; F_s]),
\end{equation}
where $F_{vl} \in \mathbb{R}^{N \times d}$ encodes both task-relevant semantics, such as scanning targets and path constraints, and external spatial context, including probe pose and environmental structure, serving as unified high-level semantic guidance for downstream fusion.

Considering that ultrasound imaging quality strongly depends on contact conditions, with pressure variations directly affecting image clarity and tissue deformation, we employ a cross-modal attention mechanism to collaboratively model ultrasound features $F_u$ and force features $F_f$.

\begin{equation}
Q_u = \mathrm{LN}(F_u)W_u^Q, \quad K_f = \mathrm{LN}(F_f)W_f^K, \quad V_f = \mathrm{LN}(F_f)W_f^V,
\end{equation}
where $\mathrm{LN}(\cdot)$ denotes Layer Normalization, and $W_u^Q, W_f^K, W_f^V$ are learnable weight matrices. 

To capture ultrasound response variations under different contact modes, including light touch, compression, and sliding, we employ a multi-head attention mechanism. Crucially, to ensure stable training and prevent the disruption of important pre-trained ultrasound features, the output projection of this cross-attention block is strictly zero-initialized.
\begin{equation}
F_{uf} = \mathrm{LN}(F_u + \mathrm{MHA}(Q_u, K_f, V_f)),
\end{equation}
where $\mathrm{MHA}(\cdot)$ denotes Multi-Head Attention. $F_{uf}$ serves as a physically meaningful local interaction representation.

Visual–language features $F_{vl}$ are further utilized to enable a unified fusion bridging local physical modeling and global semantic constraints. Through semantic-alignment-driven feature reweighting, the VLM features act as queries to adaptively select task-relevant semantic information from the force-modulated ultrasound features $F_{uf}$. Specifically,
\begin{equation}
Q_{vl} = \mathrm{LN}(F_{vl})W^Q, \quad K_{uf} = \mathrm{LN}(F_{uf})W^K, \quad V_{uf} = \mathrm{LN}(F_{uf})W^V,
\end{equation}
\begin{equation}
F_{\mathrm{fus\_attn}} = \mathrm{LN}\left( F_{vl} + \mathrm{MHA}(Q_{vl}, K_{uf}, V_{uf}) \right).
\end{equation}

The representation is subsequently enhanced through a feedforward network, which encodes semantic information, spatial context, internal anatomical structures, and physical interactions, providing a unified and high-quality conditional input for subsequent flow-matching-based action generation.

\begin{equation}
F_{\mathrm{m}} = \mathrm{LN}\left( F_{\mathrm{fus\_attn}} + \mathrm{FFN}\left( F_{\mathrm{fus\_attn}} \right) \right),
\end{equation}
where $\mathrm{FFN}(\cdot)$ denotes a two-layer multi-layer perceptron (MLP) with GELU activation. It is worth noting that the final linear projection layer of this fusion module is strictly zero-initialized. This carefully designed strategy guarantees that the module initially acts as an identity mapping, thereby preserving the representational integrity of the pre-trained action expert model during the early stages of fine-tuning.

\subsection{Stage-Adaptive Modulation Mechanism (SAMM)}

Embodied ultrasound scanning exhibits stage-dependent interactions, transitioning from non-contact to stable contact. Visual and language information dominate during non-contact, whereas force and ultrasound signals become critical after contact. Fixed-weight fusion cannot effectively capture these dynamic modality shifts and may introduce interference. We therefore propose a Stage-Adaptive Modulation Mechanism (SAMM), which explicitly models force-driven interaction stages and dynamically adjusts multimodal fusion strength.

\begin{figure*}[t]
\centering
\includegraphics[width=0.9\textwidth]{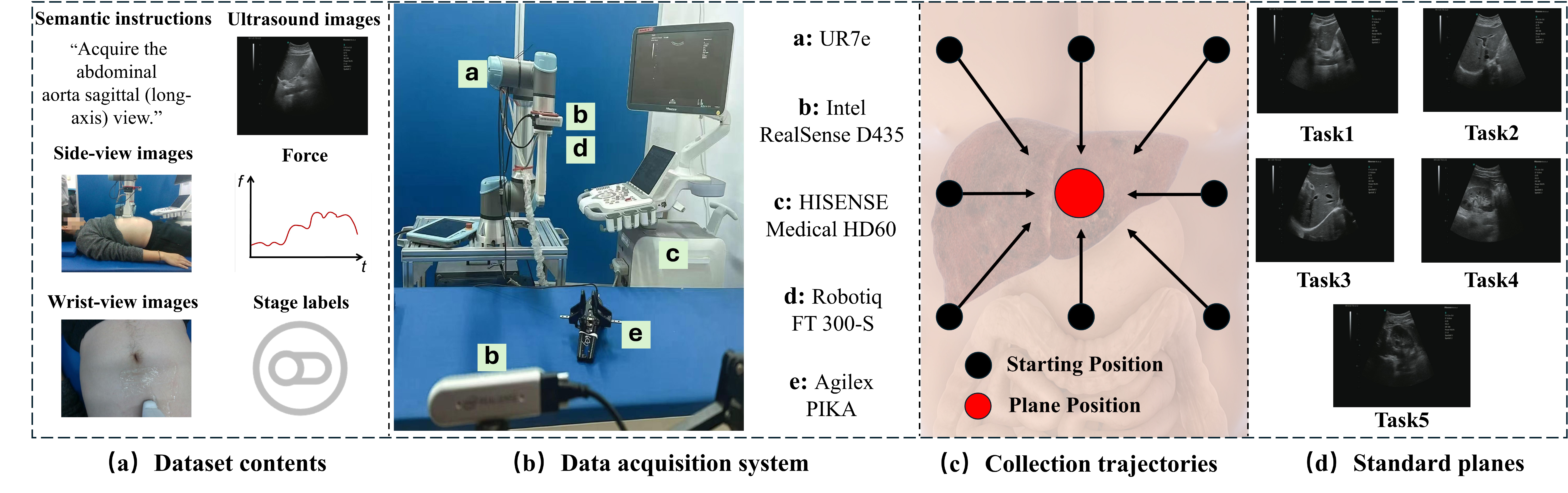}
\caption{Overview of ForceU-VLA-Data, illustrating the dataset contents, acquisition system, collection trajectories, and representative standard planes.}
\label{fig3}
\end{figure*}

The contact force $f$ measured by a six-axis force sensor is denoted as follows
\begin{equation}
f = (f_x, f_y, f_z),
\end{equation}
where $f_x$, $f_y$, and $f_z$ represent the force components along the x-, y-, and z-axes, respectively.

First, the Euclidean norm of the force is calculated to provide a continuous representation of contact strength
\begin{equation}
F = \|f\|_2 = \sqrt{f_x^2 + f_y^2 + f_z^2}.
\end{equation}

Rather than employing discrete stage divisions, we model the interaction stages with a continuous function that maps the force magnitude to a fusion modulation coefficient, enabling a smooth transition from the non-contact to the contact state. The stage-adaptive weight $\alpha(F)$ is defined as:
\begin{equation}
\alpha(F) = \alpha_{\mathrm{low}} + \left( \alpha_{\mathrm{high}} - \alpha_{\mathrm{low}} \right) \cdot \sigma\left( \frac{F - \tau}{\gamma} \right),
\end{equation}
where $\sigma(\cdot)$ denotes the Sigmoid function. The parameter $\tau$ represents a learnable force threshold that characterizes the critical point at which contact occurs, with an initial value of 5\,N.
In practice, stable contact between the probe and human soft tissue typically occurs within a force range of approximately 3--8\,N, making 5\,N a reasonable initial center. The parameter $\gamma$ controls the smoothness of the transition region and is also learnable, with an initial value set to 2. The parameters $\alpha_{\text{low}}$ and $\alpha_{\text{high}}$ correspond to the lower and upper bounds of the fusion strength for the non-contact and contact stages, respectively, set to 0.1 and 1. 

When the system is in the non-contact stage (i.e., $F \ll \tau$), the modulation coefficient satisfies $\alpha(F) \approx 0.1$, retaining only a small portion of the fused force–ultrasound features to prevent noise from adversely affecting action decisions. In the stable contact stage (i.e., $F \gg \tau$), $\alpha(F) \approx 1.0$, fully incorporating the fused features and thereby enhancing the modeling of tissue contact states and ultrasound imaging information.
Within the transition region, the weight varies continuously and differentiably, ensuring stable training of the model.

During the action sequence generation stage, the base features $F_{vl}$ that are extracted from the VLM suffix are combined with the modulated multimodal features $F_m$ to predict the velocity field of the flow-matching policy:

\begin{equation}
v_t = \mathrm{Linear}\left( F_{vl} + \alpha(F) \cdot F_m \right),
\end{equation}
where $v_t$ represents the predicted velocity field vector used in the Ordinary Differential Equation solver to generate a continuous sequence of operational commands known as action chunks.

\subsection{Overall Training Objective}
During the training stage, the dataset provides binary stage labels $s \in {0,1}$, indicating the non-contact and stable contact states for each time step, respectively.

Based on these labels, the target modulation weight is explicitly constructed as follows:
\begin{equation}
\alpha^* = 
\begin{cases} 
\alpha_{\mathrm{low}}, & s = 0 \\
\alpha_{\mathrm{high}}, & s = 1 
\end{cases}
\end{equation}

To ensure that the stage modulation accurately captures the interaction dynamics, an auxiliary supervision constraint is introduced. We employ a Mean Squared Error (MSE) loss to enhance the interpretability and stability of the stage modeling, defined as:

\begin{equation}
\mathcal{L}_{\text{stage}} = \mathbb{E}_{(F, s) \sim \mathcal{B}}\left[ \left( \alpha(F) - \alpha^* \right)^2 \right],
\end{equation}
where $\mathcal{B}$ denotes a mini-batch of training samples. 

Finally, the overall optimization objective of the model is formulated as a linear combination of the primary action prediction loss and the auxiliary stage modulation loss:
\begin{equation}
\mathcal{L} = \mathcal{L}_{\text{action}} + \lambda \mathcal{L}_{\text{stage}},
\end{equation}
where $\lambda$ is a balancing coefficient empirically set to 0.1. Through this joint optimization strategy, the model is able to not only learn precise action generation but also explicitly align force feedback with interaction stages, thereby significantly enhancing the robustness of the multimodal fusion process.

\begin{figure*}[t]
\centering
\includegraphics[width=0.95\textwidth]{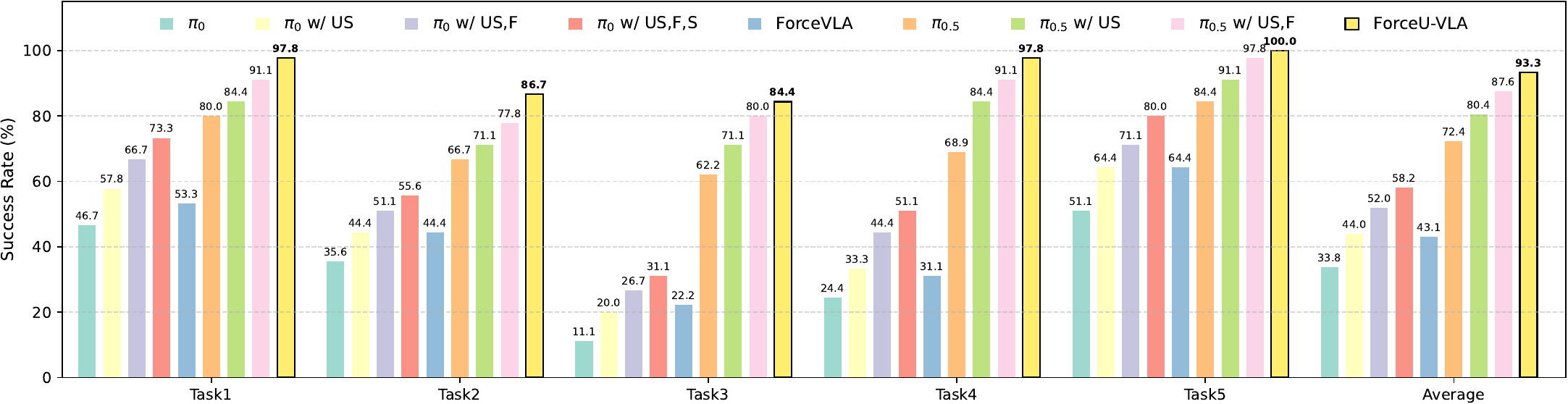}
\caption{Comparison of success rates between the proposed method and baseline approaches across five tasks. The proposed method significantly outperforms all baselines.}
\label{fig4}
\end{figure*}

\section{Dataset}
We construct a multimodal synchronized dataset, termed ForceU-VLA-Data, which integrates semantic instructions, visual information, ultrasound images, force signals, and stage labels, as illustrated in Fig.~\ref{fig3}(a). It is designed to support perception modeling and closed-loop decision-making in embodied ultrasound scanning.

As illustrated in Fig.~\ref{fig3}(b), the data acquisition system is built upon a robot-assisted ultrasound platform equipped with a UR7e six-degree-of-freedom robotic arm. Visual information is provided by two cameras: side-view images are captured by a fixed third-person camera (Intel RealSense D435, 640×480, 30 FPS), while wrist-view images are obtained from an end-effector-mounted camera (Intel RealSense D435, 640×480, 30 FPS). Ultrasound images are acquired using a HISENSE Medical HD60 system and converted into synchronized image streams via an ACASIS VC-X7PRO capture card. Force signals are recorded using a Robotiq FT 300-S six-axis force sensor. The stage labels are manually annotated during data collection to indicate the contact state between the probe and the human body (stage = 0 for no contact, stage = 1 for skin contact). As shown in Fig.~\ref{fig3}(c), to enhance data diversity and generalization, trajectories are initialized from eight different directions and collected across three body types (slim, normal, and obese). All data are acquired via teleoperation using the PIKA system, conducted by two experienced sonographers. In terms of task design, the dataset encompasses five representative clinical ultrasound scanning tasks spanning two organs (liver and the kidney). The detailed semantic instructions are defined as follows: Task 1: acquire the abdominal aorta sagittal (long-axis) view; Task 2: acquire the subxiphoid transverse view; Task 3: acquire the right subcostal oblique view through the right hepatic dome; Task 4: acquire the left kidney long-axis view; Task 5: acquire the left kidney transverse view through the renal hilum. Each task corresponds to a clinically standardized plane, as illustrated in Fig.~\ref{fig3}(d). The dataset is collected from 10 subjects, with each subject performing every task nine times. In total, ForceU-VLA-Data comprises 450 expert demonstration trajectories, amounting to approximately 100,000 high-quality synchronized timesteps. ForceVLA-Data includes five tasks and contains 244 trajectories, therefore our dataset, with a substantially larger scale, is sufficient to support effective model training.

\section{Experiments}

\subsection{Experimental Setups}

\textbf{Compared Baselines.} We compare the proposed method with a series of representative and strong baselines, including $\pi_{0}$ \cite{bib35}, $\pi_{0.5}$ \cite{bib36}, ForceVLA, and their corresponding variants (with ultrasound images (US), with ultrasound images (US) \& force signals (F), with ultrasound images (US) \& force signals \& Stage-Adaptive Modulation Mechanism (S)). Among them, $\pi_{0}$ and $\pi_{0.5}$ are flow-matching-based vision–language–action (VLA) models, pretrained on large-scale datasets and exhibiting strong generalization capabilities, flexibility, and robustness in real-world tasks. ForceVLA adopts a Mixture-of-Experts (MoE) architecture to jointly model and learn features from force and visual modalities. For fair comparison, all methods are uniformly fine-tuned on our constructed dataset before inference and evaluation.

\begin{figure*}[t]
\centering
\includegraphics[width=0.9\textwidth]{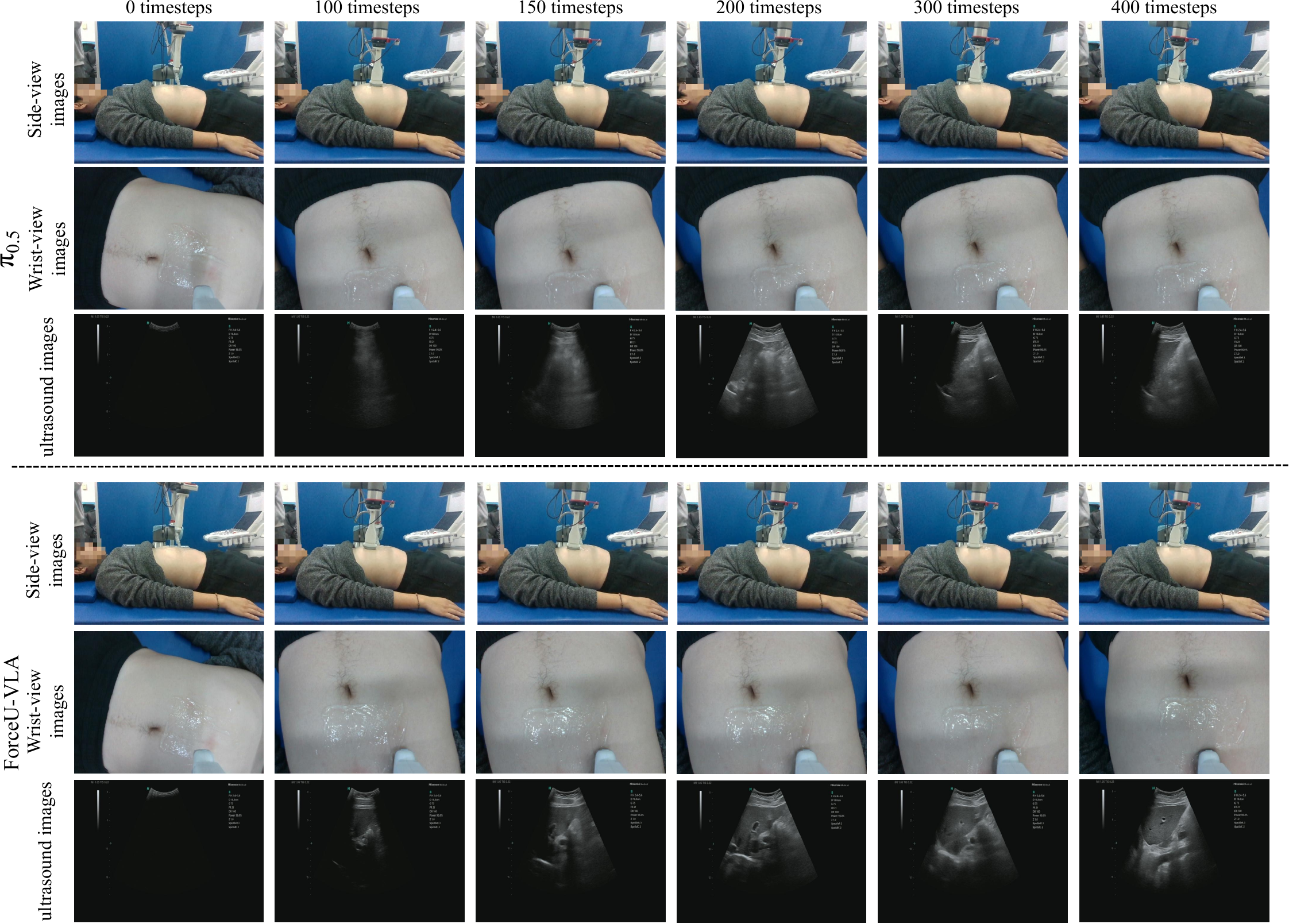}
\caption{Visualization comparison of the scanning processes of $\pi_{0.5}$ and ForceU-VLA on Task 1.}
\label{fig5}
\end{figure*}

\textbf{Evaluation Metrics.} To comprehensively evaluate different methods, we adopt success rate, timesteps, and force signals as evaluation metrics. The success rate is defined as the proportion of trials in which the target standard plane is successfully acquired, directly reflecting the task completion capability of each policy. For each task, 45 independent trials are conducted across subjects from three BMI categories (slim, normal, and obese), with 15 trials per category. Timesteps indicate task completion efficiency, while force signals characterize the physical interaction between the probe and the human body, providing an indirect assessment of scanning comfort and safety.

\textbf{Implementation Details.} The experiment is conducted on a computing platform equipped with 8 NVIDIA RTX 3090 GPUs. The model is trained for a total of 30,000 iterations, with parameters optimized using the AdamW optimizer. The learning rate is set to $5 \times 10^{-5}$. The batch size is set to 8. Both data collection and model inference are performed at a frequency of 15 Hz.

\subsection{Evaluation Results}

\textbf{Comparison of Success Rates.} Fig.~\ref{fig4} compares the success rates of ForceU-VLA and baseline methods across five tasks, averaged over three subjects. ForceU-VLA achieves the best performance on all tasks, with an average success rate of 93.3\%. Compared with ForceVLA, which obtains 43.1\%, our method improves the average success rate by 50.2\%. It also consistently outperforms $\pi_{0}$ by 35.1\%, despite the latter incorporating ultrasound, force feedback, and stage-adaptive modeling. On the more challenging Task 3, ForceU-VLA surpasses $\pi_{0.5}$ by 22.2\%, demonstrating the effectiveness of ultrasound–force fusion and stage-adaptive interaction modeling. Results for participants with different BMI categories are provided in the supplementary material.

\begin{table}[t]
\centering
\setlength{\tabcolsep}{4pt}  
\caption{Scanning timesteps on five ultrasound tasks. Bold values denote the best performance.}
\label{tab1}
\begin{tabular}{c|cccccc}
\hline
Method & Task1 & Task2 & Task3 & Task4 & Task5 & Average \\ \hline
$\pi_0$          & 593 & 831 & 611 & 557 & 399 & 598.2 \\
$\pi_0$ w/ US    & 461 & 735 & 601 & 457 & 351 & 521.0 \\
$\pi_0$ w/ US, F & 420 & 702 & 586 & 410 & 342 & 492.0 \\
$\pi_0$ w/ US, F, S & 414 & 695 & 536 & 435 & 328 & 481.6 \\
ForceVLA         & 450 & 742 & 669 & 466 & 365 & 538.4 \\
$\pi_{0.5}$      & 580 & 645 & 532 & 440 & 301 & 499.6 \\
$\pi_{0.5}$ w/ US & 423 & 576 & 476 & 351 & 287 & 422.6 \\
$\pi_{0.5}$ w/ US, F & 351 & 563 & 447 & 274 & 260 & 379.0 \\
\textbf{ForceU-VLA}          & \textbf{329} & \textbf{528} & \textbf{419} & \textbf{237} & \textbf{243} & \textbf{351.2} \\ \hline
\end{tabular}
\end{table}

\textbf{Comparison of Time Steps.} 
Table~\ref{tab1} reports the average scanning timesteps across five tasks for three subjects. The results show that the ForceU-VLA achieves the efficiency on all tasks, with an average of 351.2 timesteps.  Compared to $\pi_{0}$, ForceVLA, and $\pi_{0.5}$, the ForceU-VLA reduces the average timesteps by 41.3\%, 34.8\%, and 29.7\%, respectively, demonstrating superior policy search capability and faster convergence. This advantage primarily stems from the ForceU-VLA’s deep collaborative modeling of multimodal information, which effectively reduces redundant exploration and accelerates the localization of standard planes.

\begin{figure}[t]
\centering
\includegraphics[width=0.48\textwidth]{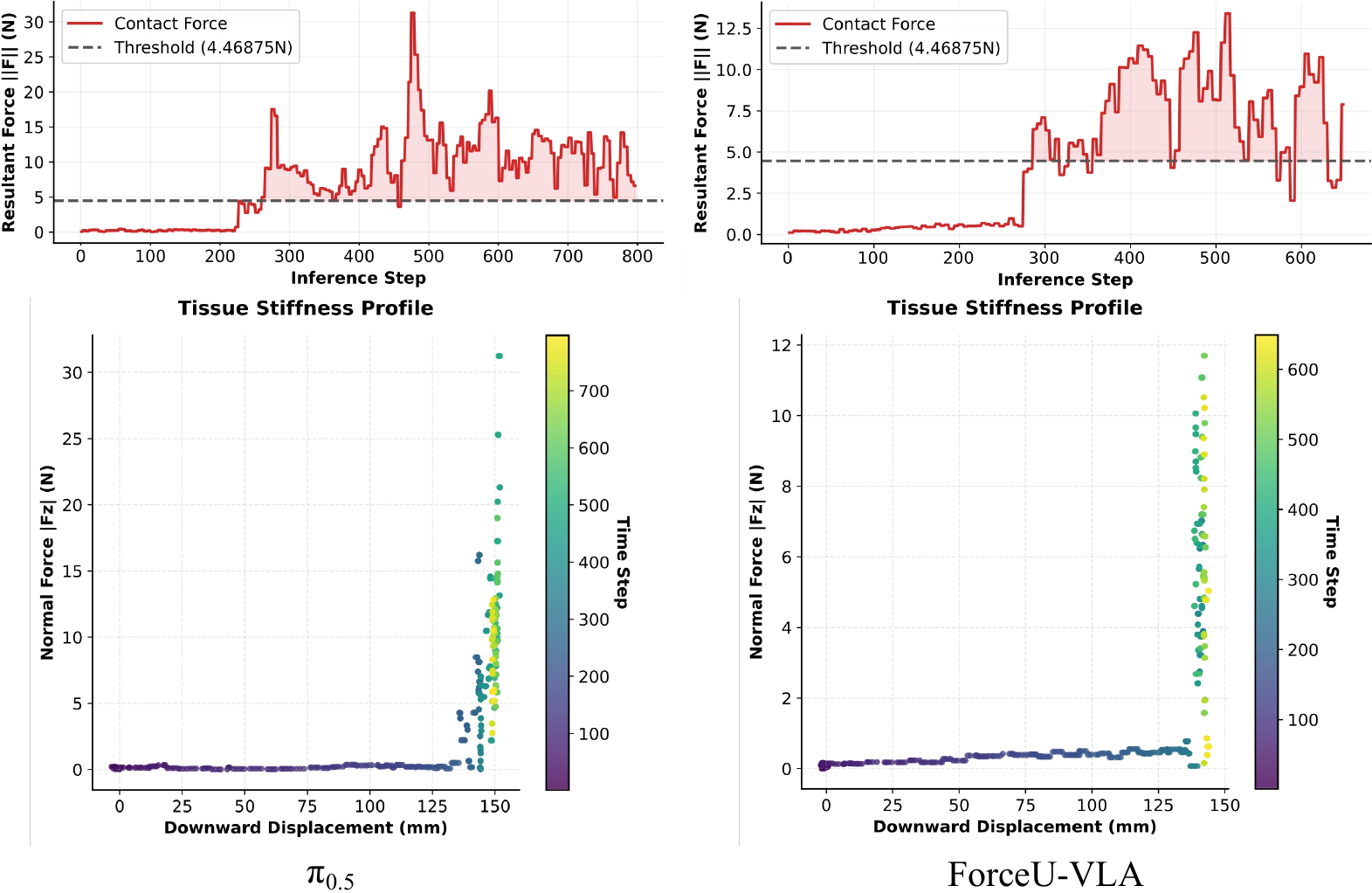}
\caption{Visualization of force comparison during the scanning processes of $\pi_{0.5}$ and ForceU-VLA on Task 2.}
\label{fig6}
\end{figure}

\begin{figure*}[t]
\centering
\includegraphics[width=0.87\textwidth]{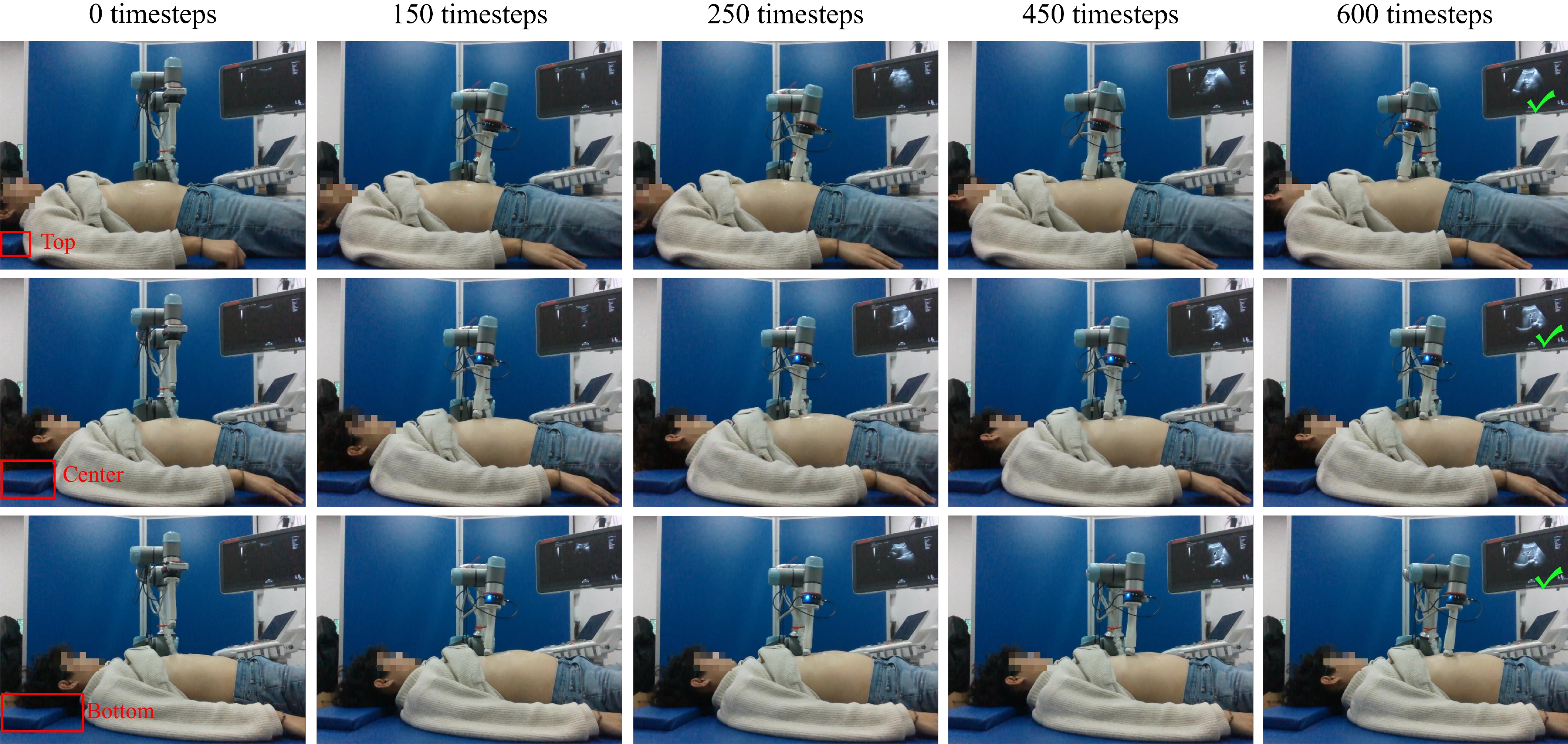}
\caption{Comparison of scanning processes under different lying positions}
\label{fig7}
\end{figure*}

\begin{table*}[t]
\centering
\setlength{\tabcolsep}{4.5pt}  
\caption{Ablation study of core Modules in ForceU-VLA.}
\label{tab2}
\begin{tabular}{lll|cccccc|cccccc}
\hline
\multicolumn{1}{c}{\multirow{2}{*}{US}} & \multicolumn{1}{c}{\multirow{2}{*}{FUSFM}} & \multicolumn{1}{c|}{\multirow{2}{*}{SAMM}} & \multicolumn{6}{c|}{Success rates $\uparrow$} & \multicolumn{6}{c}{Scanning timesteps $\downarrow$} \\ \cline{4-15} 
\multicolumn{1}{c}{} & \multicolumn{1}{c}{} & \multicolumn{1}{c|}{} & Task1 & Task2 & Task3 & Task4 & Task5 & Average & Task1 & Task2 & Task3 & Task4 & Task5 & Average \\ \hline
 & & & 80.0 & 66.7 & 62.2 & 68.9 & 84.4 & 72.4 & 580 & 645 & 532 & 440 & 301 & 499.6 \\
$\checkmark$ & & & 84.4 & 71.1 & 71.1 & 84.4 & 91.1 & 80.4 & 423 & 576 & 476 & 351 & 287 & 422.6 \\
$\checkmark$ & $\checkmark$ & & 91.1 & 77.8 & 80.0 & 91.1 & 97.8 & 87.6 & 351 & 563 & 447 & 274 & 260 & 379.0 \\
$\checkmark$ & $\checkmark$ & $\checkmark$ & 97.8 & 86.7 & 84.4 & 97.8 & 100.0 & 93.3 & 329 & 528 & 419 & 237 & 243 & 351.2 \\ \hline
\end{tabular}
\end{table*}

Furthermore, as shown in Fig.~\ref{fig5}, we visualize the scanning process of different methods. Compared with $\pi_{0.5}$, ForceU-VLA follows a more efficient search trajectory and converges faster, locating the target standard plane with fewer interaction steps. At 150 timesteps, it already captures richer ultrasound information, and at 200 and 300 timesteps, it progressively refines the probe position through fine-grained adjustments. In contrast, $\pi_{0.5}$ performs more redundant exploration and lacks effective refinement, failing to reach the standard plane even at 400 timesteps. This improvement is mainly attributed to the synergistic guidance of ultrasound and force feedback, which provides more discriminative state information at early stages, narrows the search space, and reduces unnecessary actions.

\textbf{Comparison of Force Awareness.} As shown in Fig.~\ref{fig6}, we visualize the force profiles of $\pi_{0.5}$ and ForceU-VLA in Task 2. $\pi_{0.5}$ is insensitive to physical force feedback, exhibiting large fluctuations with peaks reaching 30 N, which may cause discomfort and undermine stable imaging. In contrast, ForceU-VLA maintains the contact force at approximately 7.5 N throughout the entire scanning process, even during fine adjustments, demonstrating superior safety and control precision. Furthermore, its force distribution is notably smoother and more uniform, avoiding abrupt variations and redundant adjustments, thereby ensuring a more stable and continuous scanning process. This advantage can be attributed to explicit force modeling and multimodal synergistic fusion in ForceU-VLA, which enables more precise regulation of probe–tissue interaction and ultimately improves both scanning quality and user comfort.

\subsection{Ablation Studies}

In this section, we conduct ablation studies using $\pi_{0.5}$ as the baseline to validate the effectiveness of each component. All reported results are averaged over three subjects.

\textbf{Effectiveness of Ultrasound Imaging.} 
The introduction of ultrasound information is crucial for overall performance. As shown in Table~\ref{tab2}, incorporating ultrasound signals leads to an average improvement of 8\% in success rate, while reducing the number of timesteps by 77. This improvement can be attributed to informative real-time ultrasound feedback, which provides direct insight into internal anatomical structures and scanning quality.

\textbf{Effectiveness of FUSFM.} 
Force information also plays a crucial role in the decision-making process. As shown in Table~\ref{tab2}, after incorporating FUSFM, consistent performance improvements are observed across all tasks. The average success rate increases by 7.2\%, while the required scanning timesteps are reduced by 43.6. These results demonstrate that FUSFM enables the model to learn more effective and efficient action policies via collaborative modeling of force signals and ultrasound data.

\textbf{Effectiveness of SAMM.} 
As shown in Table~\ref{tab2}, incorporating SAMM improves the average success rate by 5.7\% and reduces scanning timesteps by 27.8. This improvement stems from SAMM’s explicit modeling of force-driven interaction stages (e.g., contact establishment and stable scanning), which enables stage-aware modulation of multimodal fusion.

\subsection{Generalization Ability}

ForceU-VLA demonstrates strong adaptability across subjects with different body types and varying lying positions. As illustrated in Fig.~\ref{fig7}, whether the subject is positioned higher or lower, the model can accurately and effectively localize the target standard plane. This robustness can be attributed to the particularly diversity and complexity of the constructed dataset, the proposed method’s strengths in multimodal collaborative modeling and decision-making, and stable performance under varying conditions, including noise, uncertainty, subtle anatomical variations, and differences in probe orientation and contact dynamics. The videos recorded from various positions are available in the supplementary materials.

\section{Conclusion} We propose ForceU-VLA, a force-aware Vision–Language–Action model that addresses the critical challenges of weak force–ultrasound coupling and the lack of stage awareness in embodied ultrasound scanning. By introducing FUSFM and SAMM, ForceU-VLA enables synergistic cross-modal interaction and stage-aware dynamic feature modulation, allowing the model to effectively capture complex probe–tissue interactions. Furthermore, we construct ForceU-VLA-Data, a real-world multimodal dataset that integrates language instructions, visual observations, ultrasound images, and force signals for embodied ultrasound scanning.

\begin{acks}
This work was supported in part by the National Science Foundation of China under Grant 62471448; in part by Shandong Provincial Natural Science Foundation under Grant ZR2024YQ004; in part by TaiShan Scholars Youth Expert Program of Shandong Province under Grant No.tsqn202312109.
\end{acks}

\bibliographystyle{ACM-Reference-Format}
\bibliography{ACM26}

\end{document}